\documentclass[11pt]{article}

\usepackage{jmlr2e}

\jmlrheading{}{2017}{}{}{}{Mozer, Kazakov, \& Lindsey}
\ShortHeadings{Discrete Event, Continuous Time RNNs}{Mozer, Kazakov, \& Lindsey}
\firstpageno{1}

\usepackage[utf8]{inputenc} 
\usepackage[T1]{fontenc}    
\usepackage{hyperref}       
\usepackage{url}            
\usepackage{booktabs}       
\usepackage{amsfonts}       
\usepackage{nicefrac}       
\usepackage{microtype}      
\usepackage{algorithm}      
\usepackage{algorithmic}      
\usepackage{listliketab}      
\usepackage{bm}      
\usepackage{amsmath}
\usepackage{makecell}
\usepackage{tabu}
\usepackage[table]{xcolor}
\usepackage{graphicx}
\usepackage{sidecap}
\RequirePackage{natbib}
\bibpunct{(}{)}{;}{a}{,}{,}

\newcommand{\sym}[1]{\text{\sffamily \scshape #1}}
\newcommand{\symx}{\text{\sffamily x}}
\newcommand{\bW}{{\ensuremath{\bm{W}}}}
\newcommand{\bU}{{\ensuremath{\bm{U}}}}
\newcommand{\bh}{{\ensuremath{\bm{h}}}}
\newcommand{\bb}{{\ensuremath{\bm{b}}}}
\newcommand{\bq}{{\ensuremath{\bm{q}}}}
\newcommand{\br}{{\ensuremath{\bm{r}}}}
\newcommand{\bs}{{\ensuremath{\bm{s}}}}
\newcommand{\bx}{{\ensuremath{\bm{x}}}}
\newcommand{\ret}{{\ensuremath{\scriptscriptstyle R}}}
\newcommand{\sto}{{\ensuremath{\scriptscriptstyle S}}}
\newcommand{\evt}{{\ensuremath{\scriptscriptstyle Q}}}
\newcommand{\tc}{{\ensuremath{\tilde{\tau}}}}
\newcommand{\btau}{{\ensuremath{\bm{\tau}}}}
\newcommand{\bM}{{\ensuremath{\hat{\bm{h}}}}}
\newcommand{\TC}{{\ensuremath{\smash{\tilde{T}}}}}

\newcommand{\wm}{{\small \sc Working memory}}
\newcommand{\cluster}{{\small \sc Cluster}}
\newcommand{\remembering}{{\small \sc Remembering}}
\newcommand{\rhythm}{{\small \sc Rhythm}}
\newcommand{\hawkes}{{\small \sc Hawkes process}}
\newcommand{\disperse}{{\small \sc Disperse}}
\newcommand{\reddit}{{\small \sc Reddit}}
\newcommand{\lastfm}{{\small \sc Last.fm}}
\newcommand{\msnbc}{{\small \sc Msnbc}} 
\newcommand{\spanish}{{\small \sc Spanish}}
\newcommand{\japanese}{{\small \sc Japanese}}

\begin{document}
\title{Discrete-Event Continuous-Time Recurrent Nets} 

%

\author{
  \name Michael C.~Mozer \email mozer@colorado.edu \\
  \addr
  Department of Computer Science\\
  University of Colorado\\
  Boulder, CO 80309
  \AND
  \name Denis Kazakov \email denis.kazakov@colorado.edu \\
  \addr Department of Computer Science\\
  University of Colorado\\
  Boulder, CO 80309
  \AND
  \name Robert V.~Lindsey \email lindsey@imagen.ai \\
  \addr Imagen Technologies\\
  New York, NY
}

\editor{}
\maketitle

\begin{abstract}

We investigate recurrent neural network architectures for event-sequence
processing. Event sequences, characterized by discrete observations stamped
with continuous-valued times of occurrence, are challenging due to the
potentially wide dynamic range of relevant time scales as well as interactions
between time scales. We describe four forms of inductive bias that should
benefit architectures for event sequences: temporal locality, position and
scale homogeneity, and scale interdependence.  We extend the popular gated
recurrent unit (GRU) architecture to incorporate these biases via intrinsic
temporal dynamics, obtaining a \emph{continuous-time GRU}.  The CT-GRU arises
by interpreting the gates of a GRU as selecting a time scale of memory, and the
CT-GRU generalizes the GRU by incorporating multiple time scales of memory and
performing context-dependent selection of time scales for information storage
and retrieval.  Event time-stamps drive decay dynamics of the CT-GRU, whereas
they serve as generic additional inputs to the GRU.  Despite the very different
manner in which the two models consider time, their performance on eleven data
sets we examined is essentially identical. Our surprising results point both to
the robustness of GRU and LSTM architectures for handling continuous time, and
to the potency of incorporating continuous dynamics into neural architectures.





\end{abstract}

\section{Introduction}

Many classic data sources in machine learning can be characterized as
sequences. For example, natural language text is a progression of words;
videos consist of a series of still images; and spoken utterances
are represented as sampled power spectra. In such sequences,
observations are ordered but there is no timing information.
In contrast to these \emph{ordinal} sequences, \emph{event} sequences consist
of observations stamped with a continuous-valued, absolute or relative time of 
occurrence.
Examples of event sequences include online product purchases, criminal activity
in a police blotter, web forum postings, an individual's restaurant
reservations, file accesses, outgoing phone calls or text messages, sent
emails, player log-ins to gaming sites, and music selections. 
In each of these examples, discrete events occur in continuous time and not
necessarily at uniform intervals.

In this article, we focus on recurrent neural net (RNN) approaches to
processing event sequences. We consider standard tasks for event sequences 
that include classification, prediction of the next event given the time lag 
from the previous event, and prediction of the time lag to the next event.

Existing sequence learning methods for recurrent nets, e.g., 
LSTM \citep{hochreiter1997} and GRU \citep{chung2014} architectures, are not
designed for event sequences but might be extended to handle them
in various ways. First, time stamps might simply be ignored
\citep[e.g.,][]{wu2016}.  Second, time might be discretized, allowing an event-based
sequence to be transformed into a sequence sampled at fixed clock intervals
\citep[e.g.,][]{hidasi2015,song2016,xwang2016,wu2017}. Third, time stamps might be
used as additional input features \citep[e.g.,][]{choi2016,du2016}.  We explore
the hypothesis that time should be handled in a more specialized manner.

To explain what we mean by `specialized,' consider the deep learning
architecture universally used for vision---the convolutional network
\citep{fukushima1980,lecun1998,mozer1987}.  Convolutional nets are successful
because they incorporate three forms of inductive bias: 
(1) \emph{spatial locality}---features at nearby locations in an image are more 
likely to have joint causes and consequences than more distant features; 
(2) \emph{spatial position homogeneity}---features deemed significant in one 
region of an image are likely to be significant in other regions; and
(3) \emph{spatial scale homogeneity}---spatial locality and position
homogeneity should apply across a range of spatial scales.

Architectures for event sequences should benefit from isomorphic forms of 
inductive bias, specifically:
(1) \emph{temporal locality}---events closer in time are more likely
to have joint causes and consequences than more distant events; 
(2) \emph{temporal position homogeneity}---event patterns deemed significant
at one point in time are likely to be significant at other points; and
(3) \emph{temporal scale homogeneity}---temporal locality and position
homogeneity should apply across a range of time scales.

We examine event-sequence architectures that incorporate these
biases, plus one more: (4) \emph{temporal scale
interactions}---sequences have different structure at different scales and
these scales interact.  Scale interactions are also found in vision, and models
have been designed to leverage these interactions by incorporating
multi-resolution pyramids at every stage of a convolutional architecture
\citep[e.g.,][]{buyssens2013,zeng2017}.  To illustrate interactions across
temporal scales, consider the scenario of online shopping. A customer may browse
various TV models one week, home automation devices
the next, and phones the next. Each of these activity
patterns indicates at least a short-term interest in some topic, but the
combination of the searches indicates a long-term interest in electronics. On
the flip side, customers may frantically shop for parts when their furnace
fails, but that does not imply a long-term interest in furnace
paraphernalia---contrary to the annoying inference that shopping sites appear to make.

\subsection{Existing approaches to event-sequence learning}

We believe that event-sequence learning can be improved because existing techniques 
fall short in incorporating the four biases we listed.  For example, one
standard technique is to include time stamps as additional inputs; these inputs
gives deep learning models
in principle all the necessary flexibility to handle time.  However, the
flexibility may simply be too great, in the same way that fully
connected deep nets are too flexible to match convolutional net performance
in vision tasks \citep{lecun1998}. The architectural biases serve to constrain 
learning in a helpful manner.

Given the similarity between the spatial regularities incorporated into
the convolutional net and the temporal regularities we described,
it seems natural to use a convolutional architecture for sequences, essentially 
remapping time into space 
\citep{waibel1990,lockett2009,nguyen2016,taylor2010,kalchbrenner2014,sainath2015,zeng2016}.
In terms of the biases that we conjecture to be helpful, convolutional nets
can check all the boxes, and some recent work has begun to investigate multiscale
convolutional nets for time series to capture scale interactions \citep{cui2016}.
However, convolutional architectures poorly address the continuous nature of
time and the potential wide range of time scales.  Consider a domain such as
network intrusion detection: event patterns of relevance can occur on a time
scale of microseconds to weeks \citep{mukherjee1994,palanivel2014}. It is
difficult to conceive how a convolutional architecture could accommodate this
dynamic range. 

RNN architectures have been proposed to address the multiscale nature of
time series and to handle interactions of temporal scale, but these approaches
have been focused on ordinal sequences and indexing is based on sequence position
rather than chronological time. This work includes clockwork RNNs
\citep{koutnik2014}, gated feedback RNNs \citep{chung2015}, and
hierarchical multiscale RNNs \citep{chungahn2016}.

%
%

A wide range of probabilistic methods have been applied to event sequences,
including hidden semi-Markov models and survival analysis
\citep{kapoor2014,kapoor2015,zhang2016}, temporal point processes
\citep{dai2016,du2015,du2016,ywang2016}, nonstationary bandits
\citep{komiyama2014}, and time-sensitive latent-factor models
\citep{koren2010}.  All probabilistic methods properly treat chronological time
as time, and therefore naturally incorporate temporal locality and position
homogeneity biases.  These methods also tend to permit a wide dynamic range of
time scales.  However, they are limited by strong generative assumptions.  Our
aim is to combine the strength of probabilistic methods---having an explicit
theory of temporal dynamics---with the strength of deep learning---having the
ability to discover representations.



\section{Continuous-time recurrent networks}

All dynamical event-sequence models must construct memories that
encapsulate information from past that is relevant for future prediction, action, 
or classification. This information may have a limited lifetime of utility, and
stale information which is no longer relevant should be forgotten.
LSTM \citep{hochreiter1997} was originally designed to operate without forgetting,
but adding a mechanism of forgetting improved the architecture \citep{gers2000}.
The intrinsic dynamics of the newer GRU \citep{chung2014} architecture incorporates
forgetting: storage of new information is balanced against the forgetting 
of old.

In this section, we summarize the GRU architecture and we characterize its
forgetting mechanism from a novel perspective that facilitates generalizing the
architecture to handling sequences in continuous time.  For exposition's sake,
we present our approach in terms of the GRU, but it could be cast in terms of
LSTM just as well.  There appears to be no functional difference between
the two architectures with proper initialization \citep{Jozefowicz2015}.

\subsection{Gated recurrent unit (GRU)}

\begin{figure}[bt]
   \begin{center}
   \includegraphics[width=4.25in]{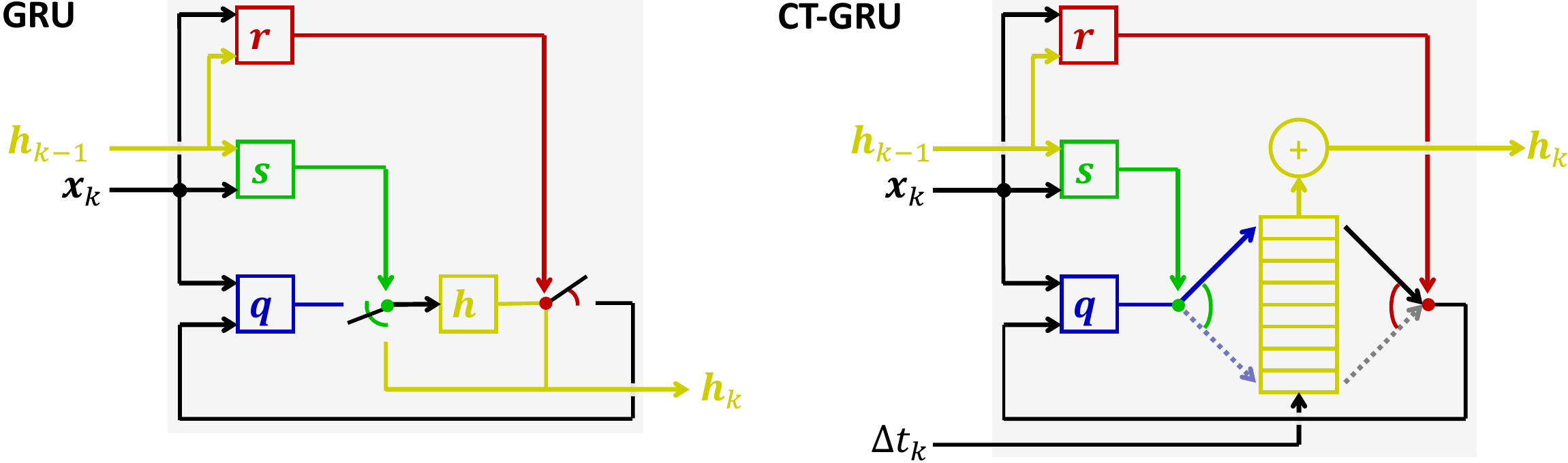}
   \end{center}
   \caption{A schematic of the GRU (left) and CT-GRU (right).
   Color coding of the elements matches the background color used in the tables
   presenting activation dynamics. For the CT-GRU, the large rectangle with
   segments represents a multiscale hidden representation. The intrinsic decay
   temporal decay of this representation, as well as the recurrent self-connections,
   is not depicted in the schematic.}
   \label{fig:CTGRU}
\end{figure}

The most basic architecture using \emph{gated-recurrent units (GRUs)} involves an 
input layer, a recurrent hidden GRU layer, and an output
layer. A schematic of the GRU units is shown in the left panel of
Figure~\ref{fig:CTGRU}.  The reset gate, $\br$, shunts the
activation of the previous hidden state, $\bh$.  The shunted state, in conjunction 
with the external input, $\bx$, is used to detect the presence of task-relevant 
events ($\bq$).  The update gate, $\bs$, then determines what proportion of the old
hidden state should be retained and what proportion of the detected event
should be stored. Formally, given an external input $\bx_k$ at step $k$ and the 
previous hidden state $\bh_{k-1}$, the GRU layer updates as follows:

\begin{center}
{\tabulinesep=.4em
\begin{tabu}{@{}|p{.43\textwidth}  p{.507\textwidth}|@{}}
   \hline
  \cellcolor{red!8}
  1. \emph{Determine reset gate settings} &
  \cellcolor{red!8}
  $\br_k \gets \mathrm{logistic} ( \bW^\ret \bx_k + \bU^\ret \bh_{k-1} + \bb^\ret )$  \\

  \cellcolor{blue!8}
  2. \emph{Detect relevant event signals} &
  \cellcolor{blue!8}
  $\bq_k \gets \mathrm{tanh} ( \bW^\evt \bx_k + \bU^\evt ( \br_k \circ \bh_{k-1}) + \bb^\evt )$  \\

  \cellcolor{green!8}
  3. \emph{Determine update gate settings} &
  \cellcolor{green!8}
  $\bs_k \gets \mathrm{logistic} ( \bW^\sto \bx_k + \bU^\sto \bh_{k-1} + \bb^\sto )$  \\

  \cellcolor{yellow!8}
  4. \emph{Update hidden state} &
  \cellcolor{yellow!8}
  $\bh_k \gets (1 - \bs_k) \circ \bh_{k-1} + \bs_k \circ \bq_k$ \\
  \hline

\end{tabu} }
\end{center}

where $\bW^*$, $\bU^*$, and $\bb^*$ are model parameters, $\circ$ denotes the Hadamard 
product, and $\bh_0 = \bm{0}$.
Readers who are familiar with GRUs may notice that our depiction of GRUs in in
Figure~\ref{fig:CTGRU} looks a bit different than the depiction in the
originating article \citep{chung2014}. Our intention is to highlight the fact
that the `update' gate is actually making a decision about what to \emph{store}
in the memory (hence the notation $\bs$), and the `reset' gate is actually
making a decision about what to \emph{retrieve} from the memory (hence the
notation $\br$). The schematic in Figure~\ref{fig:CTGRU} makes obvious the
store and retrieval operations via the gate placement on the input to and
output from the hidden state, $\bh$, respectively.

To incorporate time into the GRU, we observe that the
storage operation essentially splits each new event, $\bq_k$, into a portion
$\bs_k$ that is stored indefinitely and a portion $1-\bs_k$ that is stored for
only an infinitesimally short period of time. Similarly, the retrieval
operation reassembles a memory by taking a proportion, $\br_k$, of a
long-lasting memory---via the product $\br_k \circ \bh_{k-1}$---and a
complementary proportion, $1-\br_k$ of a very very short-term memory---a memory
so brief that it has decayed to $\bm{0}$. The retrieval operation is thus 
equivalent to computing the mixture $\br_k \circ \bh_{k-1} + 
(1-\br_k) \circ \bm{0}$.  

The essential idea of the model we will introduce, the CT-GRU, is to endow each
hidden unit with multiple \emph{memory traces} that span a range of time scales, in
contrast to the GRU which can be conceived of as having just two time scales:
one infinitely long and one infinitesimally short. We define time scale in the
standard sense of a linear time-invariant system, operating according to the
differential equation $dh/dt = -h/\tau$, where $h$ is the memory, $t$ is
continuous time, and $\tau$ is a (nonnegative) \emph{time constant} or
\emph{time scale}. These dynamics
yield exponential decay, i.e., $h(t) = e^{-t/\tau} h(0)$ and $\tau$ is the time
for the state to decay to a proportion $e^{-1} \approx .37$ of its initial level. 
The short and long time scales of the GRU correspond to the limits $\tau \to 0$ 
and $\tau \to \infty$, respectively.

\subsection{Continuous-time gated recurrent unit (CT-GRU)}


We argued that the storage (or update) gate of the GRU decides how to
distribute the memory of a new event across time scales, and the retrieval (or
reset) gate decides how to collect information previously stored across time
scales.  Binding memory operations to a time scale is sensible for any
intelligent agent because different activities require different memory
durations.  To use human cognition as an example, when you are told a phone
number, you need remember it only for a few seconds to enter it in your phone;
when making a mental shopping list, you need remember the items only until you
get to the store; but when a colleague goes on sabbatical and returns a year
later, you should still remember her name.  Although individuals typically do
not wish to forget, forgetting can be viewed as adaptive \citep{anderson1989}:
when information becomes stale or is no longer relevant, it only interferes
with ongoing processing and clutters memory.  Indeed, cognitive scientists have
shown that when an attribute must be updated frequently in memory, its current
value decays more rapidly \citep{altmann2002}.  This phenomenon is related to
the benefit of distributed practice on human knowledge retention:
when study is spaced versus massed in time, memories are more durable
\citep{mozer2009}.

Returning to the CT-GRU, our goal is to develop a model that---consistent with
the GRU---stores each new event at a time scale deemed appropriate for it, and
similarly retrieves information from an appropriate time scale.  Thus, we wish
to replace the GRU storage and retrieval \emph{gates} with storage and retrieval
\emph{scales}, computed from the external input and the current hidden state.
The scale is expressed in terms of a time constant.

To store event $k$ at an arbitrary scale $\tau_k^\sto$ (the superscript $s$
denotes `storage'), each would require a separate trace. Because separate
traces are not feasible, we propose instead a fixed set of traces with
predefined time scales, and each to-be-stored event is distributed among
the available traces. Specifically, we propose a fixed set of $M$ traces with
log-linear spaced time scales, $\TC \equiv \{ \tc_1, \tc_2, \ldots \tc_M \}$,
and we approximate the storage of a single trace at scale $\tau_k^\sto$ with a
mixture of traces from $\TC$. Of course, an exponential curve with an arbitrary
decay rate cannot necessarily be modeled as a mixture of exponentials with
predefined decay rates.  However, we can attempt to ensure that the \emph{half
life} of the mixture matches the half life of the target. Through
experimentation, we have achieved a match with high fidelity when $s_{ki}$, the
proportion of the to-be-stored signal allocated to fixed scale $i$, is:
\begin{equation}
\textstyle{s_{ki} \gets {e^{-[\ln ( \tc_i / \tau_k^\sto )]^2}}/{\sum_j e^{-[\ln ( \tc_j / \tau_k^\sto )]^2}}}.
\label{eq:ski}
\end{equation}
\begin{figure}
   \begin{center}
   \includegraphics[width=5in]{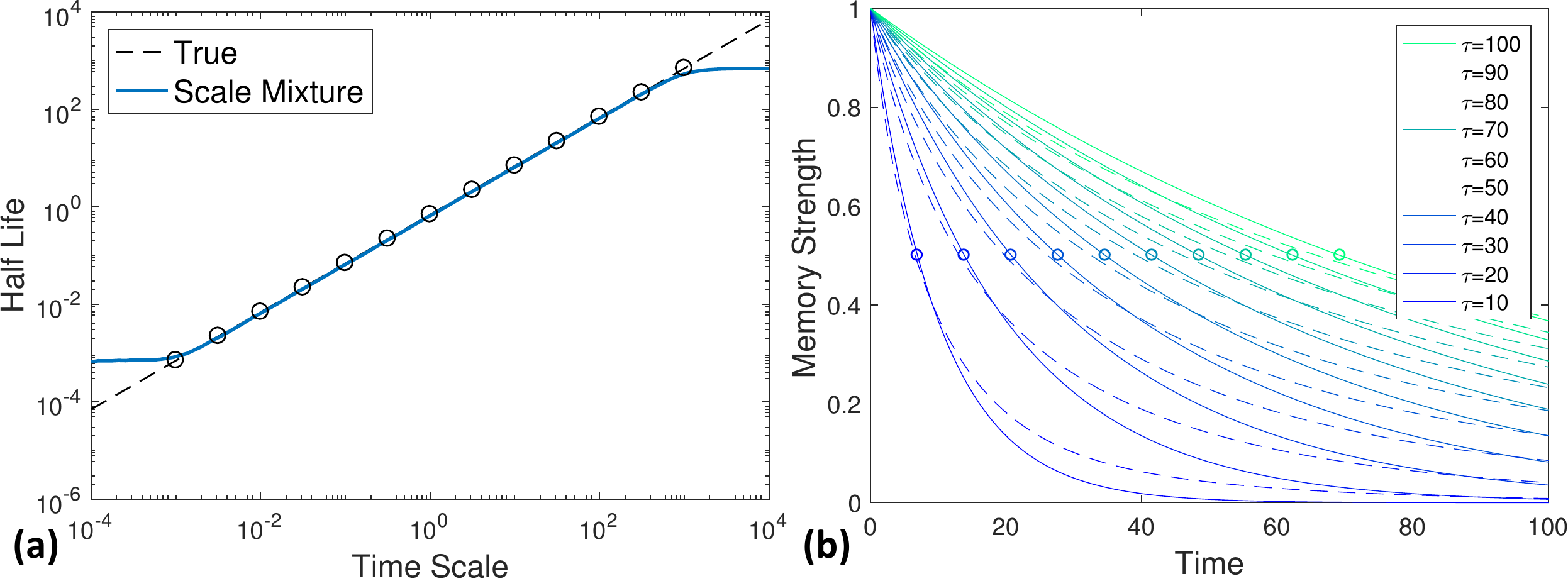}
   \end{center}
   \caption{(a) Half life for a range of time scales: true value (dashed black line)
   and mixture approximation (blue line). (b)
   Decay curves for time scales $\tau \in [10, 100]$ (solid lines)
   and the mixture approximation (dashed lines).
   \label{fig:scaleMixture}}
\end{figure}
Figure~\ref{fig:scaleMixture}a shows that the half life of arbitrary time
scales can be well approximated by a finite mixture of time scales.  The graph
plots half life as a function of time scale, with the veridical mapping shown
as a dashed black line, and the approximation shown as the solid blue line for
$\TC$ consisting of the open circles on the graph. Obviously, one cannot
extrapolate to time scales outside the range in $\TC$, but enough should be known
about a problem domain to determine a bounding range of time scales.
We have found that constraining separation among constants in $\TC$ such that 
$\tc_{i+1} = 10^{1/2}  \tc_i$ achieves a high-fidelity match.
Figure~\ref{fig:scaleMixture}b
plots memory decay as a function of time for time scales $\{ 10, 20, \ldots,
100\}$ (solid lines), along with the mixture approximation (dashed lines) using
the set of scales in Figure~\ref{fig:scaleMixture}a. Corresponding solid and
dashed curves match well in their half lives (open circles), despite the fact
that the approximation is more like a power function, with faster decay early on
and slower decay later on.  The heavy tails should, if anything, be helpful for
preserving state and error gradients.  For the sake of modeling, the important
point is that a continuous change in $\tau_k^\sto$ produces continuous
changes in both the weightings $s_{ki}$ and the effective decay function.

Just as the GRU determines the position of the storage (update) 
gate from its input, the CT-GRU determines the time scale of storage, 
$\tau_k^\sto$. We use an exponential transform to ensure nonnegative
$\tau_k^\sto$:
\begin{equation}
\btau_k^{\sto} \gets \exp \left( \bW^\sto \bx_k + \bU^\sto \bh_{k-1} + \bb^\sto \right) .
\label{eq:tauk}
\end{equation}
Functionally, the $\{ s_{ki} \}$ derived from $\tau_k^\sto$ serve as gates on each 
of the fixed-scale traces.  
The retrieval operation mirrors the storage operation. A time scale of retrieval,
$\tau_k^\ret$ is computed from the input, and a half-life-matching mixture of the 
stored traces serves as the retrieved value from the CT-GRU memory.
The right panel of Figure~\ref{fig:CTGRU} shows a schematic of the CT-GRU with
$\bs$ and $\br$ used to select the storage and retrieval scales from a
multiscale memory trace.  The time lag between events, $\Delta t_k$, is
an explicit input to the memory, used to determine the amount of decay between
discrete events.
The CT-GRU and GRU updates are composed of the identical steps,
and in fact the CT-GRU with just two scales, $\TC = \{0, \infty\}$, and fixed
$\Delta t_k$, is identical to the GRU.
The dynamics of storage and retrieval simplify because the logarithmic term in 
Equation~\ref{eq:ski} cancels with the exponentiation in Equation~\ref{eq:tauk},  
yielding the elegant CT-GRU update:
\begin{center}
{\tabulinesep=.4em
\begin{tabu}{@{}|p{.43\textwidth}  p{.507\textwidth}|@{}}
  \hline
  \cellcolor{red!8}
  1. \emph{Determine retrieval scale and weighting} &
  \cellcolor{red!8}
  \tabulinesep=0em
  \makecell[l]{
  $\ln \btau_k^\ret \gets \bW^\ret \bx_k + \bU^\ret \bh_{k-1} + \bb^\ret$ \\
  $\br_{ki} \gets \mathrm{softmax}_i \left( -(\ln \btau_k^\ret - \ln \tc_i)^2 \right)$
  } \\

  \cellcolor{blue!8}
  2. \emph{Detect relevant event signals}  &
  \cellcolor{blue!8}
  $\bq_k \gets \mathrm{tanh} ( \bW^\evt \bx_k + \bU^\evt ( \sum_i \br_{ki} \circ \bM_{k-1,i}) + \bb^\evt )$  \\

  \cellcolor{green!8}
  3. \emph{Determine storage scale and weighting} &
  \cellcolor{green!8}
  \tabulinesep=0em
  \makecell[l]{
  $\ln \btau_k^\sto \gets \bW^\sto \bx_k + \bU^\sto \bh_{k-1} + \bb^\sto$ \\
  $\bs_{ki} \gets \mathrm{softmax}_i \left( -(\ln \btau_k^\sto - \ln \tc_i)^2 \right)$
  } \\

  \cellcolor{yellow!8}
  4. \emph{Update multiscale state} &
  \cellcolor{yellow!8}
  $\bM_{ki} \gets \left[ (1 - \bs_{ki}) \circ \bM_{k-1,i} + \bs_{ki} \circ \bq_k \right] e^{- \Delta t_{k} / \tc_i} $ \\

  \cellcolor{yellow!8}
  5. \emph{Combine time scales} &
  \cellcolor{yellow!8}
  $\bh_{k} \gets \sum_i \bM_{ki}$ \\
  \hline

\end{tabu} }
\end{center}

%
%
%
%
%


\section{Experiments}

We compare the CT-GRU to a standard GRU that receives
additional real-valued $\Delta t$ inputs.  Although the CT-GRU is derived from
the GRU, the CT-GRU is wired with a specific form of continuous time dynamics,
whereas the GRU is free to use the $\Delta t$ input in an
arbitrary manner.  The conjecture that motivated our work is that the inductive
bias built into the CT-GRU would enable it to better leverage temporal
information and therefore outperform the overly flexible, poorly constrained
GRU.

We have conducted experiments on a diverse variety of event-sequence data sets,
synthetic and natural. The synthetic sets were designed to reveal the types of
temporal structure that each architecture could discover. We have explored
a range of classification and prediction tasks. The punch line of
our work is this: \emph{Although the CT-GRU and GRU handle time in very
different manners, the two architectures perform essentially identically.} We
found almost no empirical difference between the models. Where one makes errors,
the other makes the same errors.
Both models perform significantly above sensible baselines, and both models 
leverage time, albeit in a different manner. Nonetheless, we will argue that
the CT-GRU has interesting dynamics and offers lessons for future research.

\subsection{Methodology}

In all simulations, we present sequences of symbolic event labels. The input
is a one-hot representation of the current event, $x_k$.  For the CT-GRU,
$\Delta t_k$---the lag between events $k$ and $k+1$---is provided as a special
input that modulates decay (see Figure~\ref{fig:CTGRU}b).  For the
GRU, $\Delta t_{k-1}$ and $\Delta t_k$ are included as standard real-valued inputs.
The output layer representation and activation function depends on the
task.  For \emph{event-label prediction}, the task is to predict the next
event, $x_{k+1}$; the output layer is a one-hot representation with a
softmax activation function. For \emph{event-polarity prediction}, the
task is to predict a binary property of the next event.  For this task, the
output consists of one logistic unit per event label; only the event that 
actually occurs is provided a target ($0$ or $1$) value.
For \emph{classification}, the task is to map a complete sequence 
to one of two classes, $0$ or $1$, via a logistic output unit.

We constructed independent Theano \citep{theano2016} and TensorFlow
\citep{tensorflow2015} implementations as a means of verifying the code. 
For all data sets, 15\% of the training set is used as validation data
model selection, performed via early stopping and selection from a range of
hidden layer sizes. 
We assess test-set performance via three measures: accuracy,
log likelihood, and a discriminability measure, AUC \citep{green1966}. We
report accuracy because it closely mirrors log likelihood and AUC on 
all data sets, and accuracy is most intuitive.
More details of the simulation methodology and a complete description of 
data sets can be found in the Supplementary Materials.

\subsection{Discovery of temporal patterns in synthetic data}

To illustrate the operation of the CT-GRU, we devised a \wm\ task requiring 
limited-duration information storage.
The input sequence consists of commands to store,
for a duration of 1, 10, or 100 time units---specified by the commands 
\sym{s}, \sym{m}, or \sym{l}---a specific symbol---
\sym{a}, \sym{b}, or \sym{c}. The
input sequence also contains symbols \sym{a}-\sym{c} in isolation to probe
memory for whether the symbol is currently stored.  For example, 
with $\symx/t$ denoting event \symx\ at time $t$, consider the sequence:
$\{\sym{m}/0, \sym{b}/0, \sym{b}/5 \}$.  The first two
events instruct the memory to store \sym{b} for 10 time units. The third 
probes for \sym{b} at time 5, which should produce a response of 1,
whereas probes $\sym{b}/25$ or $\sym{a}/5$ should produce 0.
\begin{figure}
   \begin{center}
   \includegraphics[width=4.5in]{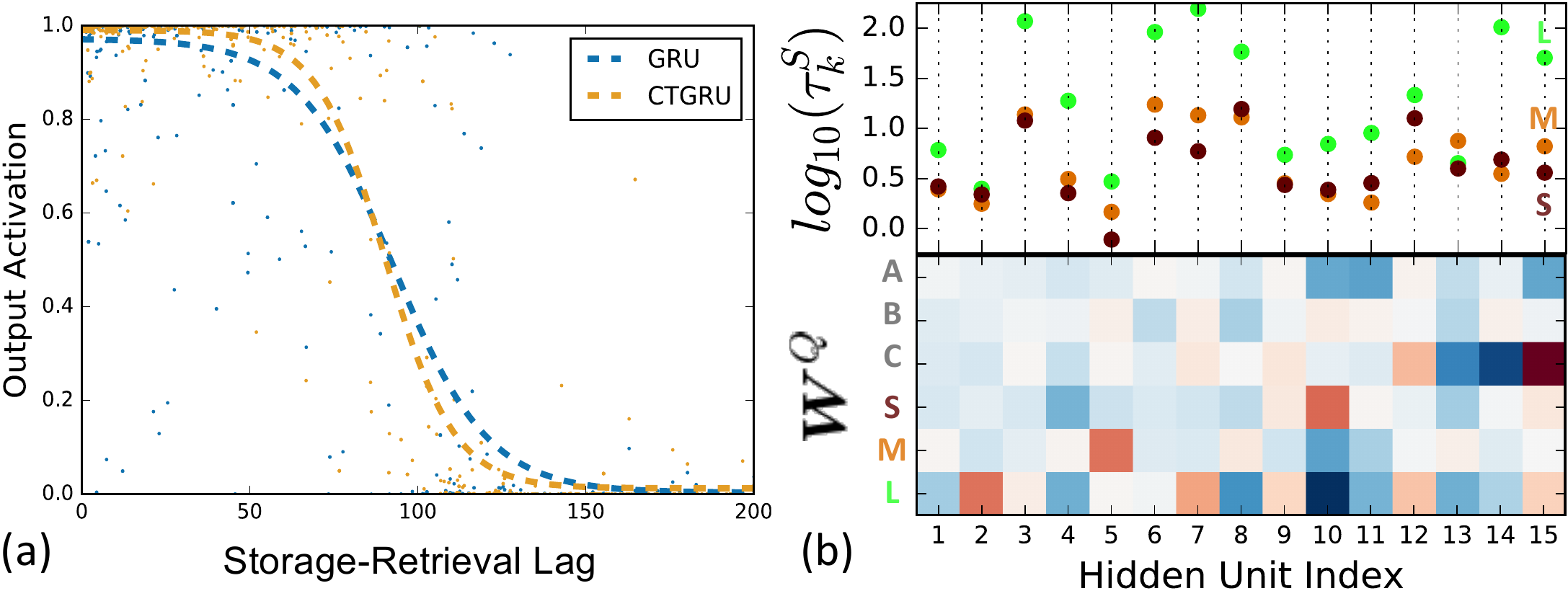}
   \end{center}
   \caption{\wm\ task: 
   (a) CT-GRU (blue) and GRU (orange) response to probe on sequences like
   $\{\sym{l}/0, \symx/0, \ldots, \symx/t \}$ for a range of  $t$.
   (b) Storage timescales, $\log_{10}(\btau_k^\sto )$, and 
   event-detection weights, $\bW^\evt$. The CT-GRU modulates
   storage time scale of symbol based on the context.
   \label{fig:denis}}
\end{figure}
Both GRU and CT-GRU with 15 hidden units learn the task well, with 98.8\% and 
98.7\% test-set accuracy, respectively. 
Figure~\ref{fig:denis}a plots probe response to sequences of the form
$\{\sym{l}/0, \symx/0, \ldots, \symx/t \}$ for various durations
$t$.  The scatterplot represents individual test sequences; the dashed line
is a logistic fit.  Both the GRU and CT-GRU show a drop off in response around
$t=100$, as desired, although the CT-GRU shows a more ideal,
sharper cut off.  Due to its explicit representation of time scale, the CT-GRU
is amenable to dissection and interpretation.  The bottom of
Figure~\ref{fig:denis}b shows weights $\bW^\evt$ for the fifteen
CT-GRU hidden units, arranged such that the units which respond more strongly
to symbols \sym{a}--\sym{c}---and thus will serve as memory for these symbols---are 
further to the right 
(blue negative, red positive). The top of the
Figure shows the storage timescale, expressed as $\log_{10} \btau_k^\sto$, for
a symbol \sym{a}--\sym{c} when preceded by commands \sym{s}, \sym{m}, or
\sym{l}. In accordance with task demands, the CT-GRU modulates the storage time
scale based on the command context.


%
\begin{figure}[b]
   \begin{center}
   \includegraphics[width=5.8in]{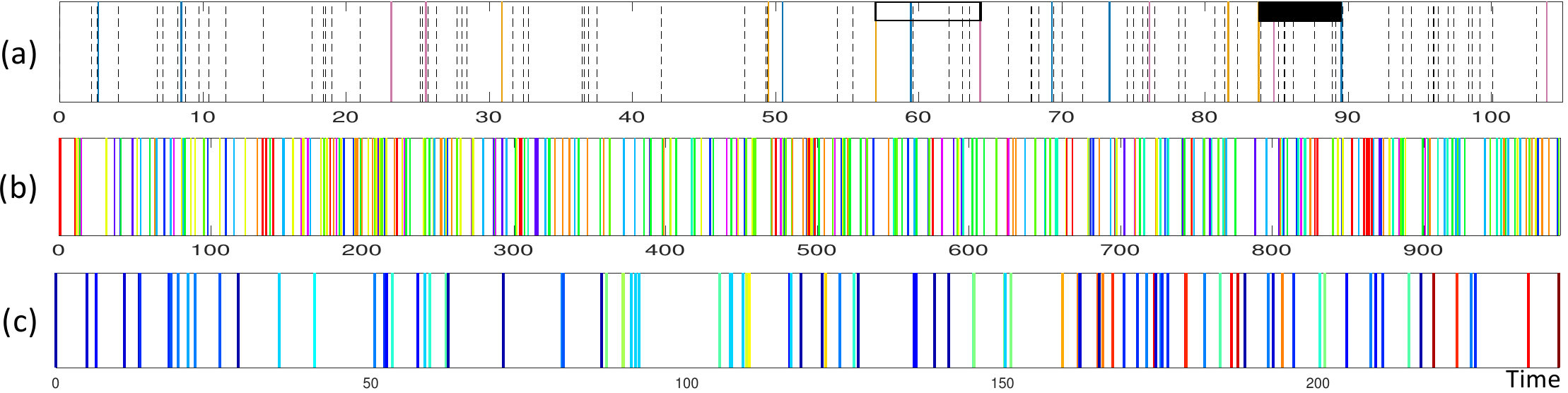}
   \end{center}
   \caption{Event sequences for (a) \cluster, 
   (b) \hawkes, and
   (c) \reddit. Time is on horizontal axis. Color denotes event label; in (a), irrelevant
   labels are rendered as dashed black lines.
   }
   \label{fig:data}
\end{figure}

\begin{figure}[bt]
   \begin{center}
   \includegraphics[width=6in]{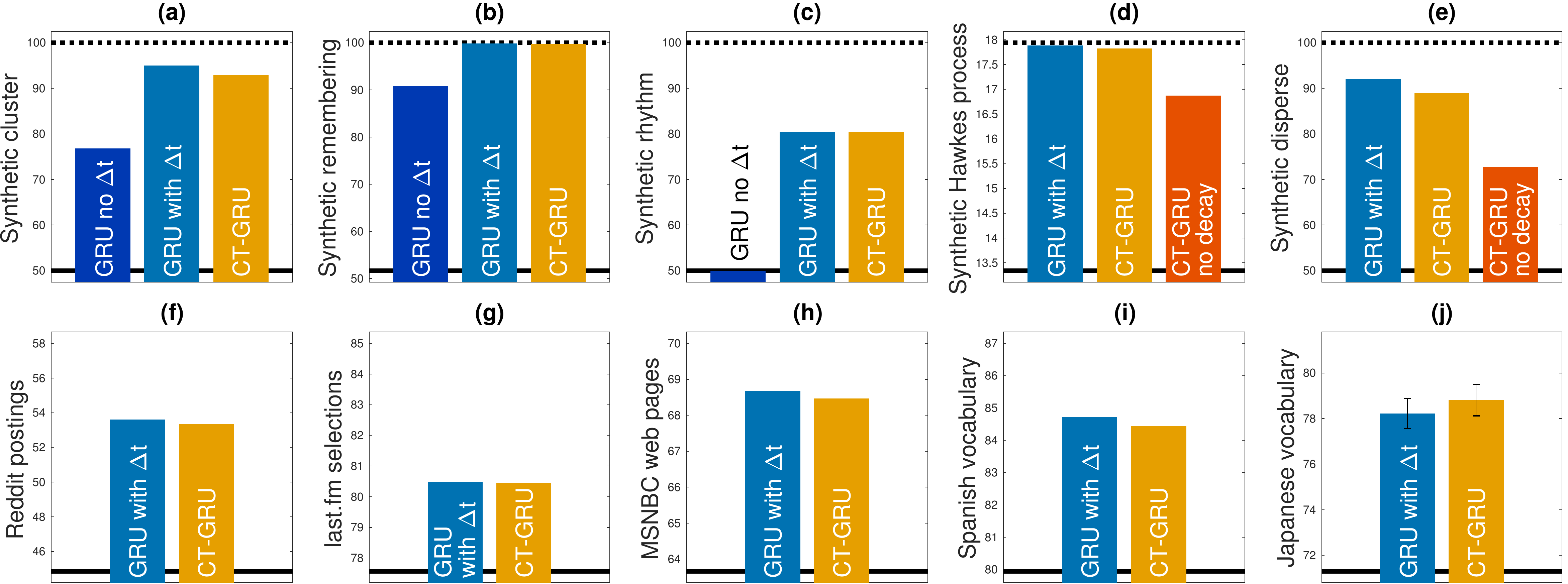}
   \end{center}
   \caption{Comparison of GRU, CT-GRU, and variants. Data sets (a)-(i) consist
   of at least 10k training and test examples and thus a single train/test split
   is adequate for evaluation. Smaller data set (j) is tested via 8-fold
   cross validation. Solid black lines represent a reference baseline
   performance level, and dashed lines indicate optimal performance (where known).}
   \label{fig:results}
\end{figure}

Moving on to more systematic investigations,
we devised three synthetic data sets for which inter-event times are required to
attain optimal performance. Data set \cluster\ classifies 100-element sequences
according to whether three specific events occur in any order
within a given time span. Figure~\ref{fig:data}a shows a sample sequence with
critical elements that both satisfy and fail to satisfy the time-span requirement,
indicated by the solid and outline rectangle, respectively.  \remembering\
outputs a binary value for each event label indicating whether the lag from the
last occurrence of the event is below or above a critical time threshold.
\rhythm\ classifies 100-element sequences according to whether the inter-event
timings follow a set of event-contingent rules, like a type of musical
notation.  The CT-GRU performs no better than the GRU with $\Delta t$ inputs,
although both outperform the GRU without $\Delta t$
(Figures~\ref{fig:results}a-c), indicating that both architectures are able to
use the temporal lags. For these and all other simulations reported, errors
produced by the CT-GRU and the GRU are almost perfectly correlated.
(Henceforth, we refer to the GRU with $\Delta t$ as the GRU.)

We ran ten replications of \cluster\ with different initializations and different
example sequences, and found no reliable difference between CT-GRU and GRU
by a two-sided Wilcoxon sign rank test ($p=.43$). Because our data sets are
almost all large---with between 10k to 100k training and test examples---and because
our aim is not to argue that the CT-GRU outperforms the GRU, we report
outcomes from a single simulation run in Figures~\ref{fig:results}a-i.

Having demonstrated that the GRU is able to leverage the $\Delta t$ inputs, we
conducted two simulations to show that the CT-GRU requires decay dynamics to
achieve its performance. We created a version of the CT-GRU in which the traces
did not decay with the passage of time. In principle, such an architecture
could be used as a flexible memory, where a unit decides which memory slot
to use for information storage and retrieval.
However, in practice removing decay dynamics
for intrinsically temporal tasks harms the CT-GRU
(Figures~\ref{fig:results}d,e).  Data set \hawkes\ consists of parallel event
streams generated by independent Hawkes processes operating over a range of
time scales; an example sequence is shown in Figure~\ref{fig:data}b.  Data set
\disperse\ classifies event streams according to whether two specific events
occur in a precise but distant temporal relationship.

\subsection{Naturalistic data sets}

We experimented with five real-world event-sequence data sets, described in detail
in the Supplementary Materials. \reddit\ is the timeseries of subreddit
postings of 30k users, with sequences spanning up to several years and a
thousand postings (Figure~\ref{fig:data}c). \lastfm\ has 300 time-tagged artist selections of 30k users,
spanning a time range from hours to months. \msnbc, from the UCI repository
\citep{lichman2013}, has the sequence of categorized MSNBC web pages viewed in
120k sessions. \spanish\ and \japanese\ are data sets of students practicing
foreign language vocabulary over a period of up to 4 months, with the lag
between practice of a vocabulary item ranging from seconds to months.  \reddit,
\lastfm, and \msnbc\ are event-label prediction tasks; \spanish\ and \japanese\ 
require event-polarity prediction (whether students successfully translated a
vocabulary item given their study history). Because students forget with the
passage of time, we expected that CT-GRU would be particularly effective for
modeling human memory strength.

Figures~\ref{fig:results}f-j reveal no meaningful performance difference
between the GRU and CT-GRU architectures, and both architectures outperform a
baseline measure (depicted as the solid black line in the Figures).  For
\reddit, \lastfm, and \msnbc, the baseline is obtained by predicting the next
event label is the same as the current label; for \spanish\ and \japanese, the
baseline is obtained by predicting the same success or failure for a vocabulary
item as on the previous trial.  Significantly beating baseline is quite
difficult for each of these tasks because they involve modeling human behavior
that is governed by many factors external to event history.

The most distressing result, which we do not show in the Figures, is that
for each of these tasks, removing the $\Delta t$ inputs from the GRU has only a
tiny impact on performance, at most a 5\% drop toward baseline. Thus, neither
GRU nor CT-GRU is able to leverage the timing information in the event stream.
One possibility is that the stochasticity of human behavior overwhelms any
signal in event timing. If so, time tags may provide more leverage for event
sequences obtained from alternative sources (e.g., computer systems, physical
processes). However, we are not hopeful given that our synthetic data sets
also failed to show an advantage for the CT-GRU, and those data sets were 
crafted to benefit an architecture like CT-GRU with intrinsic temporal dynamics.


\subsection{Summary of other investigations}

We conducted a variety of additional investigations that we summarize here.
First, we hoped that with smaller data sets, the value of the inductive
bias in the CT-GRU would give it an advantage over the GRU, but it did not.  Second,
we tested other natural and synthetic data sets, but the pattern of results is as
we report here. Third, we considered additional tasks that might reveal an advantage
of the CT-GRU such as sequence extrapolation and event-timing prediction.
And finally, we developed literally dozens of alternative
neural net architectures that, like the CT-GRU, incorporate the forms of
inductive bias described in the introduction that we expected to be helpful for
event-sequence processing. All of these architectures share intrinsic
time-based decay whose dynamics are modulated by information contained in the
event sequence. These architectures include: variants of the CT-GRU in which
the retrieved state is also used for output and computation of the storage and
retrieval scales; the LSTM analog of the CT-GRU, with multiple temporal scales;
and a variety of memory mechanisms whose internal dynamics are designed to mimic
mean-field approximations to stochastic processes, including survival processes
and self-excitatory and self-inhibitory point processes (e.g., Hawkes
processes).  Some of these models are easier to train than others, but, in the
end, none beat the performance of generic LSTM or GRU architectures provided
with additional $\Delta t$ inputs.

\section{Discussion}

Our work is premised on the hypothesis that event-sequence processing in RNN
architectures could be improved by incorporating domain-appropriate 
inductive bias. Despite a concerted, year-long effort,
we found no support for this hypothesis. 
Selling a null result is challenging. We have demonstrated that there is no
trivial or pathological explanation for the null result, such as implementation
issues with the CT-GRU or the possibility that both architectures simply ignore
time.  Our methodology is sound and careful, our simulations extensive and
thorough.  Nevertheless, negative results \emph{can} be influential, e.g., the
failure to learn long-term temporal dependencies
\citep{hochreiter2001,hochreiter1998,bengio1994,mozer1992} led to the discovery
of novel RNN architectures.  Further, this report may save others from a
duplication of effort. We also note, somewhat cynically, that a large fraction
of the novel architectures that are claimed to yield promising results one year
seem to fall by the wayside a year later.

One possible explanation for our null result may come from the fact that the
CT-GRU has no more free parameters than the GRU. In fact, the GRU has
more parameters because the inter-event times are treated as additional inputs
with associated weights in the GRU. The CT-GRU and GRU have different sorts of
flexibility via their free parameters, but perhaps the space of solutions they
can encode is roughly the same. Nonetheless, we are a bit mystified as to how
they could admit the same solution space, given the very different manners in
which they encode and utilize time.

Our work has two key insights that ought to have value for future research.
First, we cast the popular LSTM and GRU architectures in terms of time-scale
selection rather than in terms of gating information flow. Second, we show
that a simple mechanism with a finite set of time scales is capable of storing
and retrieving information from a continuous range of time scales.

To end on a more positive note, incorporating continuous-time dynamics into
neural architectures has led us to some observations worthy of further pursuit.
For example, consider the possibility of multiple events occurring
simultaneously, e.g., a stream of outgoing emails might be coded in terms of
the recipients, and a single message may be sent to multiple individuals.  The state
of an LSTM, GRU, or CT-GRU will depend on the order that the individuals are
presented. However, we can incorporate into the CT-GRU absorption time dynamics 
for an input $x$, via the closed-form solution to differential equations
$dh=-h/\tau_{hid}+x/\tau_{in}$ and $dx=-x/\tau_{in}$, yielding a model whose
dynamics are invariant to order for simultaneous events, and relatively
insensitive to order for events arriving closely in time. Such behavior could
have significant benefits for event sequences with measurement noise or random
factors influencing arrival times.




\acks{This research was supported by NSF grants DRL-1631428 and SES-1461535.}

\newpage

\appendix
\section{}

\subsection{Simulation methodology}

We constructed independent theano \citep{theano2016} and tensorflow
\citep{tensorflow2015} implementations as a means of verifying the code. For
all data sets, 15\% of the training set is used as validation data model
selection, performed via early stopping and selection from a range of hidden
layer sizes.  Optimization was performed via RMSPROP. Drop out was not used as
it appeared to have little impact on results. We assessed performance on a test
set via three measures: accuracy of prediction/classification, log likelihood
of correct prediction/classification, and AUC (a discriminability measure)
\citep{green1966}. Because accuracy mirrored the other two measures for our
data sets and because it is the most intuitive, we report accuracy.
For event-label prediction tasks, a response is correct if the highest output
probability label is the correct label. For classification tasks and
event-polarity prediction tasks, a response is correct if the error magnitude 
is less than 0.5 for outputs in $[ 0,1 ]$.

\subsubsection{GRU initialization}

The GRU $\bU^*$ and $\bW^*$ weights are initialized with $L_2$ norm 1 and such
that the fan-in weights across hidden units are mutually orthogonal.  The GRU
$\bb^*$ are initialized to zero.  Other weights, including the mapping from
input to hidden and hidden to output, are initialized by draws from a
$\mathcal{N}(0,.01)$ distribution.

\subsubsection{CT-GRU initialization}

The CT-GRU requires specifying a range of time scales in advance. These scales
are denoted $\TC \equiv \{ \tc_1, \tc_2, \ldots \tc_M \}$ in the main article.
We picked a range of scales that spanned the shortest inter-event times to
the duration of the longest event sequence, allowing information from early
in a sequence to be retained until the end of the sequence. The time constants
were chosen in steps such that $\tc_{i+1} = 10^{1/2} \tc_i$, as noted in the
main article. The range of time scales yielded $M \in \{4, 5, ..., 9\}$.
We note that domain knowledge can be useful in picking time scales to avoid
unnecessarily short and long scales.

The CT-GRU $\bU^*$, $\bW^*$, and $\bb^\evt$ parameters are initialized in the 
same manner as the GRU. The $\bb^\sto$ and $\bb^\ret$ parameters are initialized to 
$\ln (\tc_1 \tc_M)^{1/2}$ to bias the storage and retrieval scales to the
middle of the scale range.

\subsection{Data sets}

We explored a total of 11 data sets, 6 synthetic and 5 natural.

\subsubsection{Synthetic Data Sets}

All synthetic data sets consisted of 10,000 training and 10,000 testing examples.

\underline{\wm}.
We devised a simple task requiring a duration-limited or working memory. The
input sequence consists of commands to store a symbol (\sym{a}, \sym{b}, or
\sym{c}) for a short (\sym{s}), medium (\sym{m}), or long (\sym{l}) time
interval---1, 10, or 100 time units, respectively. The input sequence also
contains the symbols \sym{a}-\sym{c} in isolation to probe the memory for
whether the symbol is currently stored.  For example, consider the sequence:
$\{0,\sym{s} \}, \{0,\sym{b}\}, \{5, \sym{b}\}$, where a $\{t, \symx\}$
denotes event \sym{x} in the input sequence at time $t$. The first 2 events
instruct the memory to store \sym{b} for 10 time units. The third event probes
for \sym{b} at time 5. This probe should produce a response of 1, whereas
queries $\{25, \sym{b}\}$ or $\{5, \sym{a}\}$  should produce a response of 0.
The specific form of sequences generated consisted of two commands to store
distinct symbols, separated in time by $t_1$ units, followed by a probe of
one of the symbols following $t_2$ units. The lags $t_1$ and $t_2$ were chosen
in order to balance the training and test sets with half positive and half
negative examples.
Only fifteen hidden units were used for this task in order to interpret model
behavior.

\underline{\cluster}.
We generated sequences of 100 events drawn uniformly from 12 labels, 
\sym{a}--\sym{l}, with inter-event times drawn from an exponential distribution 
with mean 1. The task is to classify the sequence depending on the occurrence
of events \sym{a}, \sym{b}, and \sym{c} in any order within a 6 time unit window.
The data set was balanced with half positive and half negative examples.
The positive examples had one or more occurrences of the target pattern at 
a random position within the sequence.  We tested a 20 hidden unit architecture.

\underline{\remembering}.
We generated sequences of 100 events drawn uniformly from 12 labels with
inter-event time lags drawn uniformly from $\{1, 10, 100\}$. Each time a symbol
is presented, the task is to remember that symbol for 310 time steps. If the
next occurrence of the symbol is within this threshold, the target output for
that symbol should be $1$, otherwise $0$.  The threshold of 310 time steps was
chosen in order that the target outputs are roughly balanced. The target output
for the first presentation of a symbol is 0.
We tested 20 and 40 hidden unit architectures.

\underline{\rhythm}.
This classification task involved sequences of 100 symbols drawn uniformly from
\sym{a}--\sym{d} and terminated by \sym{e}.  The target output at the end of
the sequence is 1 if the sequence follows a fixed rhythmic pattern, such that
the lag following \sym{a}-\sym{d} are 1, 2, 4, and 8, respectively.
The positive sequences follow the pattern exactly. The negative sequences
double or halve between one and four of the lags. The training and test sets are
balanced between positive and negative examples. Note that this task cannot
be performed above chance without knowing the inter-event lags.
We tested 20 and 40 hidden unit architectures.

\underline{\hawkes}.
We generated interspersed event sequences for 12 labels from independent Hawkes
processes. A Hawkes process is a self-excitatory point process whose intensity
(event rate) at time $t$ depends on its history: $\lambda(t) = \mu + \alpha /
\tau \sum_{t_i < t} e^{-(t-t_i)/\tau}$, where $\{ t_i \}$ is the set of
previously generated event times.  Using the algorithm of \citep{dassios2013},
we synthesized sequences with $\alpha=.5$, $\mu=.02$, and $\tau \in \{1, 2, 4,
8, ..., 4096 \}$.  For each sequence, we assigned a random permutation of the
possible $\tau$ scales to event labels.  The intensity function ensures that
the event rate is identical across scales, but labels with shorter time
constants are more concentrated and bursty. The task here is to predict the
next event label given the time to the next event, $\delta t_k$, and the
complete event history.  Sequences ranged from 240 to 1020 events. 
Optimal performance for this data set was determined via maximum likelihood
inference on the parameters of the model that generated the data. 
We tested 10, 20, 40, and 80 hidden unit architectures.

\underline{\disperse}.
We generated sequences of 100 events drawn from 12 labels, 
\sym{a}-\sym{l}, with inter-event times drawn from an exponential distribution
with mean 1. The task is to classify a sequence according to whether
\sym{a} and \sym{b} occur separated by 10 time units anywhere in the sequence.
The target output is 1 if they occur at a lag ranging in [9,11], or 0 otherwise.
The training and test sets are balanced with half positive and half negative
examples.  We tested 20, and 40 hidden unit architectures.

\subsubsection{Naturalistic data sets}


\underline{\reddit}.
We collected sequence of subreddit postings from 30,733 users, and divided the
users into 15,000 for training and 15,733 for testing. The posting sequences
ranged from 30 subreddits to 976, with a mean length of 61.0. (We excluded users
who posted fewer than 30 times.)  Each posting was considered an event and the
task is to predict the next event label, i.e., the next subreddit to which the
user will post.  To focus on the temporal pattern of selections rather than the 
popularity of specific subreddits, we re-indexed each sequence such that each 
subreddit was mapped to the order in which it appeared in a sequence. Consequently,
the first posting for any user will correspond to label \sym{1}; the second posting
could either be a repetition of \sym{1} or a new subreddit, \sym{2}. If the
user posted to more then 50 subreddits, the 51st and beyond were assigned to
label \sym{50}.
Baseline performance is obtained by predicting that event $k+1$ will be the
same as event $k$.

\underline{\lastfm}.
We collected sequences of musical artist selections from 30,000 individuals,
split evenly into training and testing sets. We picked a span of time wide
enough to encompass exactly 300 selections.  This span ranged from under an
hour to more than six years, with a mean span of 76.3 days.  To focus on the
temporal pattern of selections rather than the popularity of specific artists,
we re-indexed each sequence such that each artist was mapped to the order in
which it appeared in a sequence. Any sequence with more than 50 distinct
artists was rejected.
Baseline performance is obtained by predicting that event $k+1$ will be the
same as event $k$.

\underline{\msnbc}.
This data set was obtained from the UCI repository and consists of the sequence
of requests a user makes for web pages on the MSNBC site. The pages are 
classified into one of 17 categories, such as \sym{frontpage}, \sym{news}, 
\sym{tech}, \sym{local}. The sequences ranged from 9 selections to 99 selections
with a mean length of 17.6. Unfortunately, time tags were not available for these
data, and thus we treated the event sequences as ordinal sequences. We were
interested in including one data set with ordinal sequences in order to examine
whether such sequences might show an advantage or disadvantage for the CT-GRU.
Baseline performance is obtained by predicting that event $k+1$ will be the
same as event $k$.

\underline{\spanish}.
This data set consists of retrieval practice trials from 180 native English
speaking students studying 221 Spanish language vocabulary items over the time
span of a semester \cite{lindsey2014}. On each trial, students were shown
an English word or phrase to translate to Spanish, and correct or incorrect
performance was recorded.  The sequences consist of a student's entire study
history for a single item, and the task is to predict trial-to-trial accuracy.
The data set consists of 37601 sequences split randomly into 18800 for training
and 18801 for testing. Sequences had a mean length of 15.9 and a maximum length
of 190.  The input consisted of $221\times2$ units each of which represents the
current trial---the Cartesian product of item practiced and incorrect/correct
performance.  The output consisted of 221 logistic units with 0/1 values for
the prediction of incorrect/correct performance on each of the 221 items.
Training and test set error is based only on the item actually practiced.
Baseline performance is obtained by predicting that the accuracy of a student's
response on trial $k+1$ is the same as on trial $k$.

\underline{\japanese}.
This data set is from a controlled laboratory study of learning Japanese vocabulary
with 32 participants studying 60 vocabulary items over an 84 day period, with
times between practice trials ranging from seconds to 50 days. For this data
set, we formed one sequence per subject; the sequences ranged from 654 to 659
trials. Because of the small number of subjects, we made an 8-fold split, each
time training on 25 subjects, validating on 3, and testing on the remaining 4.
Baseline performance is obtained by predicting that the accuracy of a student's
response on trial $k+1$ is the same as on trial $k$.

\newpage
\bibliography{references}

\end{document}